\pdfoutput=1

\documentclass[11pt]{article}
\usepackage{EMNLP2022}
\usepackage{times}
\usepackage{latexsym}
\usepackage[T1]{fontenc}
\usepackage[utf8]{inputenc}
\usepackage{booktabs}
\usepackage{microtype}
\usepackage{makecell} 

\input{webis-paper-prefix}

\newcommand{\weat}{\textsc{Weat}}

\newcommand{\static}{Static}
\newcommand{\frage}{FrecAgn}
\newcommand{\finetuned}{Fine-Tuned}
\newcommand{\decontext}{Decontext}

\newcommand{\bspubtag}{
    \vspace*{-18.6cm}\hspace*{-0.5cm}
    {\fontsize{6}{8}\selectfont%
        \renewcommand{\arraystretch}{0.9}
        \begin{tabular}{l}
            Accepted to \\
            Findings of the Association for Computational Linguistics: EMNLP 2022
        \end{tabular}
}}

\begin{document}
    \title{No Word Embedding Model Is Perfect:\\Evaluating the Representation Accuracy for Social Bias in the Media}

\author{
    Maximilian Spliethöver \textsuperscript{1}, Maximilian Keiff \textsuperscript{2}, and Henning Wachsmuth \textsuperscript{1}\\
    \textsuperscript{1}Leibniz University Hannover, Institute of Artificial Intelligence \\
    \textsuperscript{2}Universität Hamburg, Department of Informatics \\
    \small{\tt \{m.spliethoever,h.wachsmuth\}@ai.uni-hannover.de,}\\ \small{\tt maximilian.keiff@studium.uni-hamburg.de} \\
}

\maketitle

\begin{abstract}
News articles both shape and reflect public opinion across the political spectrum.
Analyzing them for social bias can thus provide valuable insights, such as prevailing stereotypes in society and the media, which are often adopted by NLP models trained on respective data.
Recent work has relied on word embedding bias measures, such as \weat{}.
However, several representation issues of embeddings can harm the measures' accuracy, including low-resource settings and token frequency differences.
In this work, we study what kind of embedding algorithm serves best to accurately measure types of social bias known to exist in US online news articles.
To cover the whole spectrum of political bias in the US, we collect 500k articles and review psychology literature with respect to expected social bias.
We then quantify social bias using \weat{} along with embedding algorithms that account for the aforementioned issues.
We compare how models trained with the algorithms on news articles represent the expected social bias.
Our results suggest that the standard way to quantify bias does not align well with knowledge from psychology.
While the proposed algorithms reduce the~gap, they still do not fully match the literature.
\end{abstract}

    \bspubtag
    \vspace*{17.8cm}\hspace*{0.5cm}

    \section{Introduction}
\label{sec:introduction}

\textit{Social bias} describes prejudices and stereotypical thinking towards certain groups in society, such as genders or ethnicities \cite{fiske1998}.
\textit{Media bias}, by contrast, refers to the tendency of media entities (e.g., a news outlet) to favor certain facts, views, or framings of events over others \cite{chen:2021}.
In this work, we focus on media bias induced by political orientations (henceforth, political bias).
While social and political bias differ in appearance, it can be expected that they relate to and mutually influence each other.
A particular political bias, \nopagebreak{for example,} may transport ideas of stereotypes, manifesting as social bias, that strengthen specific political ideas in society \cite{seiter1986,domke1999}.
Vice versa, holding particular stereotypical views may make people more susceptible to a political view promoted by a news outlet \cite{schwarz2020}.
Figure~\ref{example-texts} shows excerpts of two news articles, conveying potential gender bias.

\bsfigure{example-texts}{Excerpts of two articles, from a right and a left news outlet according to \emph{allsides.com}. Both show potential gender bias, but of different kind. The articles are included in the corpus presented in Section~\ref{sec:data}.}

The outlined kinds of bias are also relevant to NLP methods that employ news articles to train models \cite{mikolov:2013} or as a knowledge source \cite{slonim:2021}.
For example, bias present in the articles may be learned and amplified by word embeddings if not explicitly accounted for.
This impacts generalization performance negatively \cite{shah2020} and may have harmful consequences in practical applications \cite{bender:2021, joseph:2020}.
So far, one hurdle to mitigate these problems is the limited reliability of common measures of social bias present in a corpus \cite{spliethoever:2021}, stemming from embedding training algorithms not tailored to low-resource situations \cite{knoche2019,spinde:2021}.

In this paper, we investigate how to assess social bias more reliably while empirically studying the interaction of social bias and political bias in US online news outlets.
In particular, we identify \emph{low-resource settings} and \emph{token frequency differences} as two main issues with existing embedding-based bias measures.
We consider social bias towards genders, ethnicities and religions, and measure it with the widely used bias measure, \weat~\cite{caliskan:2017}.
We restrict our political bias view to the unidimensional spectrum from left to right \cite{duckitt:2010}, ignoring objectivity and fairness aspects \cite{chen2020}.

In psychology literature, stereotypical views have been shown to coincide with political orientations (Section~\ref{sec:relatedwork}), suggesting that the political views of news outlets coincide with social biases.
Under this premise, we aim to find out what word embedding algorithm best serves to reliably measure social bias.
We investigate weaknesses of a standard algorithm that stem from the reliance on word lists, infrequent tokens in the data, and the quality of embeddings.
We suggest (1)~training frequency agnostic embeddings to compensate for lower quality of rare tokens, (2)~a fine-tuned language model to account for smaller datasets, and (3)~decontextualized embeddings to alleviate the ``unnatural input'' problem with contextualized models.

For our experiments, we introduce a large-scale media bias corpus in Section~\ref{sec:data}, covering more than 500,000 news articles from 47 English-language US online news outlets over 12 years (2010--2021).
Given the corpus, we evaluate each potential improvement and compare their capability to encode and represent social bias a text corpus (Section~\ref{sec:experiments}).
To this end, we systematically generate word embedding models from subsets of different political biases.
In a second analysis, we explore the development of social bias in outlets over time in a respective manner.
We can quantify the considered types of social bias for all models using \weat.

Our results in Section~\ref{sec:experiments} provide evidence that the general embeddings quality improves notably over standard static embeddings.
Additionally, the proposed algorithms better model the expected social bias, though still not fully align with the literature.

This work provides three contributions to computational research on bias in language:
\begin{enumerate}
\setlength{\itemsep}{0pt}
\item
Findings on how to combine embedding models and bias measures to adequately quantify social bias in text corpora;
\item
a large-scale news resource annotated for political bias; and
\item
empirical insights into the interaction of social and media bias in US online news, and its development over time.%
\footnote{Code and data at \href{https://github.com/webis-de/EMNLP-22}{github.com/webis-de/EMNLP-22}.}
\end{enumerate}

    \section{Related work}
\label{sec:relatedwork}

We consider social bias that manifests as stereotypes, that is, generalized beliefs about social outgroups based on experiences with single members \cite{fiske1998}.
Such beliefs may lead to prejudices and discrimination that cause lasting harm.
Stereotypes are usually transported through language, uttered either implicitly or explicitly \cite{wodak:2008}.
If entities with high public outreach, such as politicians and media outlets, spread stereotypes, this may therefore profoundly impact their audiences \cite{seiter1986,domke1999}.

Psychology and political science literature study the relation of stereotypes with political aspects.
As part of this, multiple layers of partisan biases have been evaluated \cite{hayes:2011,bauer:2015,clifford:2020}.
Focusing on social values, \citet{valentino:2005} find a general shift of public social values related to a shift in voting outcomes.
Other works compare social values of conservatives and liberals: While liberals seem more likely to reject ``ingroup values'', conservatives emphasize tradition and religion \cite{sylwester:2015}.
Accordingly, \citet{webster:2014} observe a higher level of self-reported prejudices towards social groups that ``challenge or violate traditional social values'' among conservative probands.
\citet{chirumbolo:2016}, finally, report that liberals tend to value social equality, whereas conservatives justify social inequality with ``the preservation of status quo''.
We use these connections between political bias and social values as a reference for our analyses.

Media bias can be evaluated from many angles, too.
For instance, \citet{chen2020} explore media bias in political news, automatically detecting incomplete reporting and evaluating its linguistic manifestations.
One of their results is that words expressing negative emotions are most correlated with selective biases.
\citet{kenix:2019}, in turn, focus on conservative and liberal news articles from the US, Australia, and the UK to understand the construction of media frames.
They find that the report framing of specific outlets aligns with their political bias.
Rather than unfairness or issue perception, our work targets the interaction of media bias with social bias in news articles.

In particular, we quantify social bias in word embeddings with the widely used \emph{Word Embedding Association Test}, \weat{} \cite{caliskan:2017}.
\weat's main idea is to calculate the cumulative distance between groups of word vectors that describe a social group and attributes.
Similar measures exist, such as \textsc{Ect} \cite{dev:2019}, \textsc{Rnsb} \cite{sweeney:2019}, and \textsc{Mac} \cite{manzini:2019}, \textsc{Ripa} \cite{ethayarajh:2019}, \textsc{WeatVec} \cite{knoche2019}, the Smoothed First-Order Co-occurrence \cite{rekabsaz:2021} and \textsc{Same} \cite{schroder:2021}
but our goal is not to find the best measure.
Rather, we seek to learn how measures like \weat{} behave for different embedding algorithms.
We are not aware of works that have done similar.

Similar to the analysis we carry out, \citet{garg:2018} exploit the properties of word embeddings to evaluate temporal relationships between changes of social bias and empirical demographic changes in the US.
They evaluate embedding models trained on texts from different decades, for example finding that gender bias decreased with the women's movement in the 1960's.
In a comparable analysis, \citet{rios:2020} find that gender bias reduced in biomedical research over time for some areas, but not in others.
In this work, we utilize \weat{} to evaluate social bias in news articles.
Unlike previous work, however, we compare word embedding algorithms to model social bias in texts and their alignment with the literature reviewed above.

Closest to our work is the research of \citet{knoche2019} and \citet{spinde:2021}.
The former use \weat{} to compare social biases present word embeddings trained on different ideological online wikis.
All wikis are found to have similar biases for gender, race, and religion, but to varying degrees.
\citet{spinde:2021} collect US news articles from a liberal and a conservative media outlet.
By training one embedding model for each outlet and measuring the differences of all words in the embedding spaces, they determine the most biased words.
The underlying hypothesis is that words, for which the context varies more strongly, will also be more biased.
We apply a data collection method similar to \citet{spinde:2021}, but cover 47 outlets.
Additionally, instead of just focusing on two extreme communities, our corpus spans a wider spectrum of political opinions.
Our main goal is to deepen the understanding of the social bias in word embeddings for different training algorithms.
    \section{Method}
\label{sec:approach}

This paper studies how to best evaluate a text corpus for social bias, harnessing the ability of word embeddings to encode direct contexts.
In particular, we quantify the social bias encoded in models trained on a corpus.
The models are thus used as a proxy from which we derive the social bias in the original corpus.
In the following, we present our evaluation method, discuss potential issues, and describe the employed embedding algorithms.

\subsection{Evaluating Social Bias in Embeddings}

We seek to analyze to what extent word embedding models encode the social bias of training data.
For further insights, we investigate the models quality.

\paragraph{Word Similarity}

The quality of the semantic space of word embedding models benefits from larger datasets \cite{pennington:2014}.
Since most social bias measures rely on this space, better embeddings should also yield more accurate bias evaluations.
To gain a better understanding of the quality, we conduct word-similarity evaluations \cite{spinde:2021} of all models we explore.
These evaluations are based on a list of word pairs, human-annotated for similarity.
For each pair, the cosine similarity between the vectors generated by a model is computed.
The Spearman's~$\rho$ between the vector similarities and the annotations represents the score.
While this intrinsic evaluation is not able to predict the performance on downstream tasks, it provides insights into the semantic quality of the embeddings \cite{faruqui:2016}.
The results also enable us put the social bias evaluation into context.
We apply two tests, \emph{MEN} \cite{bruni:2014} and \emph{WordSim353} \cite{finkelstein:2001}.%
\footnote{\href{https://github.com/EloiZ/embedding_evaluation}{https://github.com/EloiZ/embedding\_evaluation}}

\paragraph{Social Bias}

To quantify social bias, we report results of \weat{} \cite{caliskan:2017}.
At its core, \weat{} relies on four word lists describing a concept.
Two lists describe social groups that are evaluated in the context of attributes which represent the other two lists.
Common combinations are:
\begin{itemize}
\setlength{\itemsep}{0pt}
    \item \emph{Gender.} Male/female and career/family terms
    \item \emph{Ethnicity.} African-/European-American names and pleasant/unpleasant terms
    \item \emph{Religion.} Christianity/Islam terms and pleasant/unpleasant terms
\end{itemize}

Using a given embedding model, all words are transformed into word vectors, in order to measure the cumulative distance between the vectors.
Let $G$ and $\tilde{G}$ be the word embeddings for the two social group lists, and $A$ and $B$ those for the attribute lists.
Now, let $\Delta(\mathbf{w}, G, \tilde{G})$ be the mean difference between the cosine similarity of a word embedding~$\mathbf{w}$ to all word embeddings in $G$ and to the embeddings in $\tilde{G}$.
Then, the \weat{} score is defined as the effect size of the difference between $A$ and $B$:
\[
    \frac
        {\text{mean}_{\mathbf{a} \in A} \Delta(\mathbf{a}, G, \tilde{G}) \;-\; \text{mean}_{\mathbf{b} \in B} \Delta(\mathbf{b}, G, \tilde{G})}
        {\text{std\_dev}_{\mathbf{w} \in A \cup B} \Delta(\mathbf{w}, G, \tilde{G})}
\]
This results in a value from --2 to 2, where 0 represents the least possible bias.
Using \weat{} makes our results comparable with related work.
We calculate \weat{} scores using the implementation of the WEFE framework \cite{badilla:2020} and use word lists of \citet{spliethoever:2021}.

\paragraph{Accuracy of Bias Evaluation}

Evaluating social bias in a word embedding model assumes that its semantic space is meaningful.
As different word embedding algorithms achieve this with varying success, they likely also differ in their accuracy in encoding social bias.
Assuming that the bias measure at hand (here, \weat) works as intended, it is possible to evaluate differences between algorithms, given a corpus with known social bias.
Below, we thus compare models of different embedding algorithms on training data for which the social bias is known from literature (see Section~\ref{sec:relatedwork}).
While we cannot derive exact \weat{} values for a corpus, we can infer relative differences for liberal and conservative texts.
Together with the results of the word similarity evaluation, we can draw conclusions regarding the reliability of the results.

\subsection{Potential Evaluation Issues}
\label{sec:approach:eval-issues}

As previous research \cite{spliethoever:2020,spinde:2021} points out, evaluating text corpora for social bias with static word embeddings (e.g., word2vec) entails three main problems:
\begin{enumerate}
\setlength{\itemsep}{0pt}
    \item \emph{Limited Corpus Size.} The training data influences the semantic quality of the embeddings.
    \item \emph{Representation Degeneration.} Token frequency differences in the training data entail embeddings of differing quality.
    \item \emph{Out-of-Vocabulary Tokens.} Limited vocabularies cause unknown tokens during evaluation.
\end{enumerate}

In the following, we describe these issues in more detail. To alleviate them, we train word embedding models with different algorithms below.

\paragraph{Limited Corpus Size}

To generate a meaningful semantic space based on context, word embedding models tend to require large datasets.
For example, the pre-trained word2vec model \cite{mikolov:2013} was trained on 100B tokens, the largest GloVe model \cite{pennington:2014} on 840B.
Thus, the quality of the embedding may suffer from small corpora.
In turn, the results of the bias evaluation may not be as accurate as with larger corpora.

\paragraph{Representation Degeneration}

Representation degeneration describes the dependence of meaningful embeddings on the token occurrences, reflecting its available number of contexts \cite{karampatsis:2020}.
It implies that infrequent tokens (\textit{rare tokens}) tend to have lower-quality embeddings than more frequent ones (\textit{popular tokens}).
While fluctuations are expected due to Zipf's law, they result in less reliable semantic encodings \cite{gong:2018,karampatsis:2020,wolfe:2021}.
In the context of social bias measures, this issue is especially relevant, since they implicitly assume a similar quality for all word vectors.
The difference between tokens can be high for certain corpora \cite{spliethoever:2020}.
Even more problematic, the occurrences also tend to vary within a single test (e.g., more male term occurrences than female ones), potentially influencing the social bias measure results negatively.

While a frequency difference can itself be a form of social bias, it makes the evaluation less straightforward, which is why we ideally seek to abstract from it.
A na\"ive way would be to artificially augment the data by duplicating contexts of rare tokens.
As we intend to keep the original signals, though, we explore more direct means of abstraction.

\paragraph{Out-Of-Vocabulary Tokens}

Static word embedding models have a fixed vocabulary, determined by tokens in their training corpus and are unable to generate embeddings for tokens not included (henceforth, OOV tokens).
However, most embedding bias measures rely on pre-defined word lists and assume that an embedding is available for each word.
OOV tokens hence need to be ignored in the evaluation, reducing the comparability of multiple models.
This can be alleviated by sub-word tokenization, as used for BERT \cite{devlin:2019}.

\subsection{Word Embedding Algorithms}

We hypothesize that no existing word embedding algorithm is able to account for all issues discussed.
Therefore, we train models with multiple algorithms, and we evaluate them against each other.
For implementation details on the different algorithms, see Appendix~\ref{sec:appendix:model-training}.

\paragraph{Static}

As baseline, we train static embedding models with word2vec \cite{mikolov:2013}.
An advantage of this method is the fast training process.
Also, static word embeddings are by now well researched and interpretable \cite{bommasani:2020}.
In turn, the algorithms require large training data to generate high-quality embeddings.
Furthermore, due to the representation degeneration problem, the measured bias may be less comparable if the token frequencies vary strongly between two corpora.
The static models will be referred to as \static{} in the following.

\paragraph{Frequency-Agnostic}

Frequency-agnostic word embeddings (FRAGE) \cite{gong:2018} aim to approach the representation degeneration problem by accounting for the frequency of tokens.
FRAGE does so by training a long short-term memory model (LSTM) on a language modeling task and introducing an adversarial discriminator, classifying tokens as rare or popular.
During training, the LSTM tries to minimize the ability of the adversarial to predict the class of each token.
While reducing the impact of token frequency, the model is trained from scratch, increasing training time requiring much data to obtain high-quality embeddings.
The models will be referred to as \frage.

\paragraph{Fine-Tuned}

To account for the shortcomings of FRAGE, we additionally fine-tune BERT.
On the one hand, it provides a good basis for embeddings, as it is pre-trained on large corpora.
This should offer a certain level of base quality for semantic embeddings, potentially reducing the negative effect of size differences in the fine-tuning data.
Moreover, it may minimize quality differences between embeddings of rare and popular tokens.
Due to sub-word tokenization, OOV tokens are also not an issue.
However, BERT contextualizes embeddings dynamically during generation, requiring the context of a token (e.g., the sentence it appears in) as input.
Since bias measures usually work with single token embeddings, we need to generate embeddings by querying the model for unnatural inputs (e.g., inputs containing only the token in question without context) \cite{bommasani:2020}.
The resulting models will be referred to as \finetuned.\,\,

\paragraph{Decontextualized}

As an alternative to fine-tuned BERT, we employ the averaged pooling strategy presented by \citet{bommasani:2020} to generate decontextualized embeddings.
The general idea is to embed all contexts of a specific token in a \textit{context dataset} using a language model.
To receive a single embedding per token, the contextualized embeddings are then averaged.
Since the final embeddings are contextualized by the context dataset, they can also be expected to encode its social bias.
We thus use the corpus we aim to evaluate for social bias as context.
Since this method is also based on BERT, we expect the embeddings to have similar advantages over static embeddings, while accounting for the unnatural-input problem.
Moreover, since the resulting embeddings are static rather than contextualized, they should retain benefits such as better interpretability.
The time needed to generate decontextualized word embeddings is, however, more dependent on the size of the context dataset, since all contexts need to be embedded separately.
This results in a potentially long generation time.
The models will be referred to as \decontext.

    \section{Data}
\label{sec:data}

\begin{table*}[t]
    \small
    \centering
    \setlength{\tabcolsep}{2.5pt}
    \begin{tabular}{lrrrrrrrrrrrrr@{\quad}r}
        \toprule
        \textbf{Orientation}  & \textbf{2010} & \textbf{2011} & \textbf{2012} & \textbf{2013} & \textbf{2014} & \textbf{2015} & \textbf{2016} & \textbf{2017} & \textbf{2018} & \textbf{2019} & \textbf{2020} & \textbf{2021} & \textbf{No Date} & \textbf{All} \\

        \midrule

        Liberal       & 4559 & 7953 & 13969 & 13474 & 21685 & 26238 & 22900 & 21302 & 17641 & 16542 & 20613 & 27114 & 71408 & 285398 \\
        \,\,\,\, \gr Left & \gr 3955 & \gr 5847 & \gr 9679 & \gr 9133 & \gr 17186 & \gr 22018 & \gr 20767 & \gr 19301 & \gr 15962 & \gr 14892 & \gr 18494 & \gr 22193 & \gr 24043 & \gr 203470 \\
        \,\,\,\, \gr Lean-left & \gr 604 & \gr 2106 & \gr 4290 & \gr 4341 & \gr 4499 & \gr 4220 & \gr 2133 & \gr 2001 & \gr 1679 & \gr 1650 & \gr 2119 & \gr 4921 & \gr 47365 & \gr 81928 \\

        \addlinespace

        Neutral     & 4800 & 3100 & 5023 & 3584 & 6304 & 7558 & 7832 & 6045 & 7299 & 8756 & 11621 & 10218 & 7535 & 89675 \\

        \addlinespace

        Conservative      & 3878 & 4746 & 6624 & 6666 & 7259 & 7922 & 8438 & 8715 & 10867 & 10224 & 14797 & 27326 & 28263 & 145725 \\
        \,\,\,\, \gr Lean-right & \gr 2392 & \gr 2521 & \gr 4083 & \gr 3203 & \gr 3617 & \gr 3659 &\gr  2715 & \gr 3814 & \gr 5259 & \gr 4434 & \gr 7583 & \gr 15382 & \gr 16298 & \gr 74960 \\
        \,\,\,\, \gr Right & \gr 1486 & \gr 2225 & \gr 2541 & \gr 3463 & \gr 3642 & \gr 4263 & \gr 5723 & \gr 4901 & \gr 5608 & \gr 5790 & \gr 7214 & \gr 11944 & \gr 11965 & \gr 70765 \\
        \midrule
Total         & {13237} & {15799} & {25616} & {23724} & {35248} & {41718} & {39170} & {36062} & {35807} & {35522} & {47031} & {64658} & {107206} & {520798} \\
        \bottomrule
    \end{tabular}
    \caption{Number of news articles per year for each orientation in our corpus (liberal, neutral, conservative) and their for sub-groups (e.g., left). The total number of articles (\emph{All}) includes those for which \emph {no date} could be extracted.}
    \label{tab:dataset-statistics}
\end{table*}

We now present the large-scale corpus that we acquired to study the existence of social bias in news articles across the political spectrum in the US.

\paragraph{Source Data}

Using media bias ratings from news aggregation platform \emph{allsides.com}, we collected articles from liberal (left and lean-left labels) and conservative (right and lean-right labels), as well as neutral (center label) outlets.
While this uni-dimensional view on the political spectrum is limited \cite{duckitt:2010}, it provides us with a clear distinction and makes results easier to interpret.
We refer to news articles with liberal, neutral, and conservative labels as \textit{data subsets} in Section~\ref{sec:experiments}.

Similar to \citet{spinde:2021}, we collected news articles from \emph{Common Crawl}%
\footnote{Common Crawl, \href{https://commoncrawl.org}{https://commoncrawl.org}}.
Since the media bias rating history is not available, we mapped each outlet to its current rating.
To extract the pure text from the collected files in WARC format, we used the library \emph{news-please} \cite{hamborg:2017}.
For our experiments, we also extracted the articles' date of publication automatically as far as possible.

\paragraph{Preprocessing}

To filter out non-English articles, we classified the language of each text automatically using the \textit{langdetect} library%
\footnote{\href{https://github.com/Mimino666/langdetect}{https://github.com/Mimino666/langdetect}}.
In contrast, we intentionally did not filter news categories (e.g., keeping only news articles about politics), in order to avoid selection bias.
Furthermore, the different embedding algorithms require varying preprocessing steps.
For word2vec, sentence splitting is required.
In order to train the FRAGE model, we tokenized the data and replaced ultra-rare tokens with ``<unk>'', since the model expects the preprocessing of the WikiText-2 corpus \cite{merity:2016}.
To do so, we used the \emph{huggingface} tokenizer%
\footnote{\href{https://github.com/huggingface/tokenizers}{https://github.com/huggingface/tokenizers}}
and ended up with a vocabulary of around 39k tokens.

\paragraph{Statistics}

In total, we collected 520,798 news articles from 47 different outlets, 19 of which are liberal, 10 neutral, and 18 conservative.
Table \ref{tab:dataset-statistics} reports detailed dataset statistics, showing that the number of articles is increasing over time, more or less monotonously.
For about 20\% of all articles (107,206), no publication date could be extracted.

    \section{Experiments}
\label{sec:experiments}

\begin{table}[t]
    \small
    \renewcommand{\arraystretch}{1}
    \centering
    \setlength{\tabcolsep}{1.85pt}
    \begin{tabular}{lrrrcrrr}
        \toprule
         & \multicolumn{3}{c}{\textbf{WordSim353}} && \multicolumn{3}{c}{\textbf{MEN}} \\
        \cmidrule(l@{2pt}r@{2pt}){2-4}  \cmidrule(l@{2pt}r@{2pt}){6-8}
        \textbf{Algorithm} & \textbf{Liberal} & \textbf{Neutr.} & \textbf{Cons.} && \textbf{Liberal} & \textbf{Neutr.} & \textbf{Cons.} \\
        \midrule
        \static & --0.02 & 0.05 & 0.07 && 0.04 & --0.01 & --0.03 \\
        \frage & 0.57 & 0.56 & 0.55 && 0.55 & 0.46 & 0.51 \\
        \finetuned & 0.25 & 0.48 & 0.30 && 0.34 & 0.46 & 0.30 \\
        \decontext & \textbf{0.64} & \textbf{0.62} & \textbf{0.65} && \textbf{0.77} & \textbf{0.74} & \textbf{0.76} \\
        \cmidrule(l@{2pt}r@{2pt}){2-4}  \cmidrule(l@{2pt}r@{2pt}){6-8}
        BERT & \multicolumn{3}{c}{0.25} && \multicolumn{3}{c}{0.21} \\
        \bottomrule
    \end{tabular}
    \caption{Spearman's~$\rho$ of the word embedding similarity evaluation on the two tests, \textit{WordSim353} and \textit{MEN}. The embedding models were trained using the evaluated algorithms on \textit{liberal}, \textit{neutral} or \textit{conservative} articles. Bold values indicate the best score in each column. For comparison, the values of pre-trained BERT are shown.}
    \label{tab:results-similarity-evaluation}
\end{table}

We now describe our experiments to evaluate embedding algorithms regarding their capabilities to accurately represent social bias in text corpora.
To do so, we assess an algorithm's ability to generate a meaningful embedding space and to avoid the issues detailed in Section~\ref{sec:approach} arising from sparse~data.

In particular, we systematically train models on all news articles with either political bias from our corpus (Section~\ref{sec:data}), once with each of the four word embedding algorithms from Section \ref{sec:approach}.
To increase the data available for each bias, we aggregate news articles for lean-left and left as \emph{liberal} as well as for lean-right and right outlets as \emph{conservative}.

\begin{table*}[t]
    \small
    \renewcommand{\arraystretch}{1}
    \centering
    \setlength{\tabcolsep}{3.5pt}
    \begin{tabular}{lrrrrcrrrrcrrrr}
        \toprule
         & \multicolumn{4}{c}{\textbf{Gender}} && \multicolumn{4}{c}{\textbf{Ethnicity}} && \multicolumn{4}{c}{\textbf{Religion}} \\
         \cmidrule(l@{2pt}r@{2pt}){2-5}  \cmidrule(l@{2pt}r@{2pt}){7-10}  \cmidrule(l@{2pt}r@{2pt}){12-15}

        {\textbf{Algorithm}} & {\textbf{Liberal}} & {\textbf{Neutr.}} & {\textbf{Cons.}} & {\textbf{$\Delta$}} && {\textbf{Liberal}} & {\textbf{Neutr.}} & {\textbf{Cons.}} & {\textbf{$\Delta$}} && {\textbf{Liberal}} & {\textbf{Neutr.}} & {\textbf{Cons.}} & {\textbf{$\Delta$}} \\

        \midrule

        \static & --0.151 & --0.169 & 0.230 & 0.381 && 0.060 & --0.061 & 0.098 & 0.038 && 0.301 & --0.002 & --0.298 & --0.600 \\
        \frage & 0.632 & 0.611 & 0.763 & 0.131 && 0.555 & 0.680 & 0.658 & \textbf{0.103} && 1.166 & 0.795 & 1.181 & 0.015 \\
        \finetuned & 0.275 & 0.036 & 0.671 & \textbf{0.396} && 0.600 & 0.659 & 0.419 & --0.182 && 0.873 & 1.235 & 0.442 & --0.431 \\
        \decontext & 0.334 & 0.409 & 0.370 & 0.036 && 0.419 & 0.422 & 0.429 & 0.010 && 0.479 & 0.486 & 0.519 & \textbf{0.040} \\

         \cmidrule(l@{2pt}r@{2pt}){2-5}  \cmidrule(l@{2pt}r@{2pt}){7-10}  \cmidrule(l@{2pt}r@{2pt}){12-15}
        BERT & \multicolumn{4}{c}{0.098} && \multicolumn{4}{c}{1.234} && \multicolumn{4}{c}{0.621} \\

        \bottomrule
    \end{tabular}
    \caption{\weat{} values of the models trained with each evaluated algorithm for the three types of social bias.~$\Delta$ denotes the difference between the values of the models trained on \textit{conservative} and \textit{liberal} articles respectively; the highest $\Delta$ for each bias type is marked bold. For reference, the \weat{} values of pre-trained {BERT} are shown.}
    \label{tab:results-weat-evaluation}
\end{table*}

\subsection{Word Similarity Tests}

To better understand the models' quality, we first evaluate their performance on word-similarity tests.

Table~\ref{tab:results-similarity-evaluation} indicates that all proposed algorithms produce more meaningful embedding spaces compared to the \static{} models. The scores of the latter are close to 0.00, suggesting little to no correlation with the actual word similarities.
A potential reason for the low scores is the limited training data, as discussed in Section~\ref{sec:approach:eval-issues}, which may not be large enough to train high-quality models from scratch. The \decontext{} models that are pretrained on a larger dataset, on the other hand, achieve the highest scores for all data subsets on both tests (ranging from 0.62 to 0.77), also notably outperforming the underlying BERT model. The fine-tuning process of \finetuned{} only marginally improves upon the base model. Considering that the liberal and conservative data subsets are notably larger than the neutral subset, it also seems that more data hurts the \finetuned{} performance. This might be an issue of over-fitting to the fine-tuning data, decreasing the applicability of the resulting embeddings for the general similarity task.
Further, the ``unnatural'' input used to generate \finetuned{} models, compared to the averaging strategy of the \decontext{} models, potentially impacts the embedding quality \cite{bommasani:2020}.

These results suggest that, while the size of the training corpus does have an impact on the quality of the word embeddings, it is not the only contributing factor.
For example, comparing the results in Table~\ref{tab:results-similarity-evaluation} across algorithms for the same dataset, the choice of the algorithm seems to be important as well.
That said, some algorithms do seem to benefit from the additional data.
While the models trained on the liberal data perform slightly better on MEN tests compared to the other two models trained on smaller data, the benefit seems to be mostly negligible considering the increase in data needed (the liberal dataset contains nearly twice as many articles compared to the conservative dataset) and the additional training time.
Furthermore, it is unclear, if this performance difference might partially also due to the selection of tested words in the respective word similarity tests.

Considering consistency, the frequency-agnostic and the decontextualized model appear most stable across all tests and data subsets.
As a result, the models are also more comparable in the \weat{} evaluation across data subsets, as the quality of the embedding models seems to be less dependent on the corpus size and content.

Overall, the suggested algorithms seem to improve the quality of the embedding space and abstract reasonably from the corpus size.
For \decontext{} and \finetuned{}, OOV tokens are less of a problem, as they train on sub-word tokens.
The impact of fewer OOV tokens seems small in Table~\ref{tab:results-weat-evaluation} than previously assumed.
The performance of the \frage{} model does not vary notably from the two models trained on sub-word tokens.
Less OOV tokens should, however, result in more accurate social bias evaluations as more word embeddings exit, from which associations can be measured.
As noted before, this can be more important when testing smaller datasets, as done in Section~\ref{sec:experiments-temporal}.

To analyze the representation degeneration, we repeated the evaluation with token pairs for which at least one was among the 100 least used tokens of the respective data subset. In general, results were similar to those in Table~\ref{tab:results-similarity-evaluation}, indicating that \decontext{} and \frage{} also perform well with rare tokens. While the results seem convincing, they must be interpreted with care. The similarity evaluations test word embedding models for general words and meaning rather than for social biases. Furthermore, the relation between these tests and the social bias measures is not fully clear.

\subsection{Bias Representation Accuracy}
\label{sec:results-weat}

As detailed in Section \ref{sec:approach}, each model is evaluated for social bias using \weat{}.
Following \citet{caliskan:2017}, positive values indicate potential biases towards women compared to men (\textit{Gender}), African-American compared to European-American names (\textit{Ethnicity}), and Islam compared to Christianity (\textit{Religion}).
Based on our literature review presented in Section \ref{sec:relatedwork}, we expect the liberal models to be biased against men, European-American names and Christianity, which should be reflected in positive \weat{} values.
Accordingly, we expect the opposite for the conservative models, and the neutral models should receive \weat{} values located between the others.

Table \ref{tab:results-weat-evaluation} shows the results.
The $\Delta$ columns indicate the difference in \weat{} values between the models trained on conservative and on liberal news articles.
It is a rough measure of an algorithm's accuracy in encoding social bias.
The closer $\Delta$ is to the maximum ($2 -(-2) = 4$), the better the models represent the expected social bias detailed in Section \ref{sec:approach}.
Our discussion relies on this relative measure, as the exact \weat{} value of the data subsets is unknown.
We chose not to use absolute values, as a negative $\Delta$ highlights cases that contradict our initial expectations, providing additional information on the quality of the word embedding models and applied measures.
Since \finetuned{} and \decontext{} are based on BERT, we report BERT's \weat{} values for reference.

For all three evaluated bias types, at least one of the suggested algorithms receives a better accuracy than \static.
While the fine-tuned models achieve the highest $\Delta$ in the gender bias evaluation, \frage{} performs closest to expectation in the ethnicity bias evaluation.
For the religion bias evaluation, \decontext{} shows the highest $\Delta$, even though the absolute differences are comparatively small.
It is noteworthy that models trained with \static{} achieve the second-best accuracy for the gender and ethnicity bias evaluations.

In general, the \weat{} values for liberal and conservative models are less divergent than expected. Also, $\Delta$ is consistently close to 0 for \decontext{} and \frage{}. When comparing the models for a single data subset (e.g., for liberal outlets only), the \weat{} value strongly depends on the applied algorithm.
The variance for the same data subset across all evaluations is in all cases above 0.7, with an average of 1.005. This is an intriguing finding, indicating that the choice of a particular algorithm is an important parameter when interpreting \weat{} results, making exact \weat{} values less meaningful and relative comparisons to a reference necessary.

A further interesting result is the fact that the \finetuned{} and \decontext{} models have lower \weat{} values than the BERT model they are based on.
We hypothesize that this is due to our data being less biased, which changed the word associations during the fine-tuning and decontextualization.
With the analysis at hand, however, this phenomenon can not be explained conclusively.
While there does not seem to be one ``best'' algorithm for evaluating social bias according to Table \ref{tab:results-weat-evaluation}, the combination of data, algorithm, and bias type seem to matter for the final result.
A potential explanation is that the social bias present in the data is not as we hypothesized in Section \ref{sec:relatedwork}, and the political bias does not correlate with social bias to the expected degree.
Neither psychology literature nor our manual inspection of samples of the corpus make this seem likely, though.

Stereotypes, ideas of society, and  with that social bias rather may be expressed more implicitly (see the example in Figure \ref{example-texts}), potentially drawing word list-based measures to quantify bias ineffective.
Similarly, the word lists applied by the \weat{} evaluations might not be fully applicable to the evaluated datasets, requiring adaptation to the given linguistic style \cite{chaloner:2019}.
For example, while liberal media may use the term ``immigrant'' to describe people coming to the US from a different country, conservative media may rather use the term ``alien'' \cite{webson:2020}.
If a word list only includes one of the terms, it cannot properly reflect the associations with the target group and thus the social bias in the data.
We suspect that both issues might contribute to the negative $\Delta$ values presented in Table~\ref{tab:results-weat-evaluation}.
In this regard, future work may investigate measures that do not rely on predefined word lists, but adapt to the corpus being evaluated.

\subsection{Temporal Evaluation}
\label{sec:experiments-temporal}

In our final experiment, we evaluated the change of social bias over time for the three political bias subsets.
This also allows for insights into how the presented algorithms work with even fewer data, similar to the analyses of \citet{garg:2018} and \citet{rios:2020}.
In particular, we trained one word embedding model for articles of each year from each political bias considered (liberal, neutral, and conservative).
We excluded all 107,206 articles for which we could not extract dates automatically.
Here, we only used the decontextualization algorithm, given that it produced meaningful embeddings across all data subsets above.
This is an important property for this evaluation, as the year-based sub-corpora are comparatively small.
We evaluate the models for social bias using \weat.

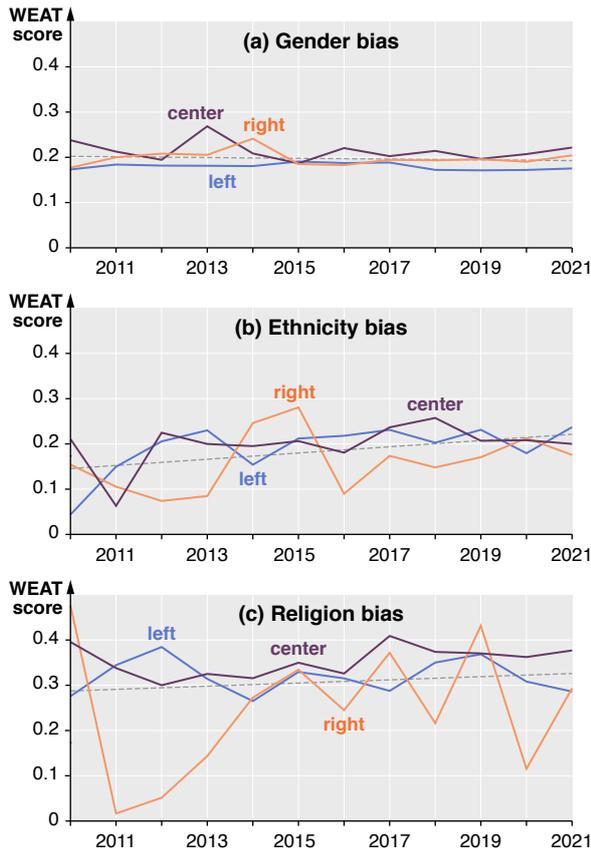
\bsfigure{temporal-all-weat}{Plots of the development of the \weat{} scores of the word embedding models for each bias type over time. Each model was trained on data subsets for each pair of year and political orientation. Gender bias slightly reduces over time, while ethnicity bias and religious increase (dashed regression lines).}

Figure \ref{temporal-all-weat} plots the results for each type of bias.
While gender bias doesn't change notably, ethnicity and religious bias increase over the 12 years.
We don't attribute this is to the amount of data, as the fluctuations of the neutral model happen mostly during years where the number of articles is similar to the conservative bias (see Table \ref{tab:dataset-statistics}).
Similar to the evaluation of the full models, we find that the relative social bias levels do represent the expected results to a certain degree.
The liberal model generally shows lower \weat{} values in the gender and religious evaluation compared to the conservative model.
For the ethnicity evaluation, the liberal and conservative models are less distinctive though and show a very similar trend.

Similar to the analysis of the full models, the small differences in \weat{} values, compared to the full \weat{} scale, might indicate that the absolute \weat{} numbers are less meaningful and only work in relative comparisons.

    \section{Conclusion}
\label{sec:conclusion}

In this paper, we have compared word embedding algorithms for the task of evaluating text corpora for social bias. To this end, we have introduced a US online news corpus that covers three political bias directions at five levels. Our literature review has motivated that specific political bias coincides with social bias with respect to gender, ethnicity, and religion. We have taken advantage of this property to train three word embedding algorithms and evaluate them for social bias using \weat{}. Lastly, we present an example application, analyzing the development of social bias in news articles over a 12 year period.

We find that the particularly frequency-agnostic and decontextualized embedding spaces are more meaningful and encode the social bias more accurately than word2vec.
They fail, however, to do so consistently for all bias types.
While the respective algorithms should be more reliable, especially when evaluating sparse datasets, the exact \weat{} results should be considered with care.
The values do not seem to quantify social bias in the same way for all embedding algorithms.
Future research should investigate the relation between \weat{} values of an algorithm and the encoded bias.

Our findings give insights into the role of word embedding algorithms within the social bias evaluation of texts, and they demonstrate what type of embedding models work even in sparse data scenarios. Thereby, we contribute to understanding social bias in texts and NLP applications in general.
    \section*{Acknowledgments}
We thank the anonymous reviewers for their valuable feedback. Furthermore, we thank the team behind \href{https://allsides.com}{https://allsides.com} for providing us with data files of their media bias ratings.
    \section*{Limitations}

One limitation of our evaluation is the distantly supervised approach used to label articles for political bias based on the outlet it was published by. We recognize that not all articles of an outlet are necessarily politically biased in the same way and to the same degree. Similarly, the political bias of an outlet could have changed over the evaluated period. A more refined approach could label articles based on their content, rather than the publishing outlet. Similar can be said for the social bias labels. Ultimately, it is not guaranteed that the social bias present in the analyzed 500k news articles statistically matches knowledge from psychology literature. Under the premise that literature is right, however, we are convinced that our inference from political to social bias to be sound, even if may not apply to the same extent to all articles.

We also acknowledge that we did not account for the completeness of the word lists used in the \weat{} evaluations, which might therefore suffer from selection bias, hence not comprehensively representing the target groups. As the \weat{} values depend on the contents of the word lists, the presented values might therefore not be fully accurate.
A potential improvement to account for representation issues is to adapt the word lists to the language of each data subset, since outlets might use different vocabularies to describe the same groups.

Lastly, we were only able to evaluate a limited number of word embedding algorithms that account for token frequency issues. Potential alternatives include KAFE \cite{ashfaq:2022}, which relies on a knowledge graph to improve token representations, and AGG \cite{yu:2022}, for which the code was not available at the time of conducting the experiments. Similarly, we chose to fine-tune our BERT model for four epochs in all cases to obtain a comparable setting. Other choices might yield varying results.
    \section*{Ethical Statement}

We generate word embedding models for encoding social bias, as we train explicitly on texts that we expect to be biased. The models might therefore also contain more bias than other pre-trained models. They were, however, solely trained for the purpose of analyzing the training data. Due to the nature of the corpus and the comparatively sparse training data, we believe that the resulting models are not very applicable to other tasks.

We also note that, as already mentioned in the limitations, the word lists that we used in the \weat{} evaluation are not complete. They might therefore not represent the social groups to a satisfying degree for real-world applications.

    \bibliography{emnlp22-social-media-bias-lit}
    \bibliographystyle{acl_natbib}

    \appendix

\section{Model training implementation details}
\label{sec:appendix:model-training}

\paragraph{Static}

We train static embedding models with the gensim implementation of the word2vec algorithm and trained them using the skip-gram method with a window size of five for five epochs.%
\footnote{\href{https://github.com/RaRe-Technologies/gensim}{https://github.com/RaRe-Technologies/gensim}}
We stick to the commonly used vector size of 300 dimensions.

\paragraph{Frequency-Agnostic}

To train frequency-agnostic models with the FRAGE algorithm, we used the AWD-LSTM implementation published by \citet{gong:2018}. For efficiency reasons, we decrease the number of epochs from 4000 to 500 and increased the batch size from 80 to 600.

\paragraph{Fine-Tuned}

For the fine-tuned language models, we chose uncased BERT as starting point. We fine-tune the model for each political bias for four epochs with a standard masked language modeling objective using the Transformers library%
\footnote{\href{https://github.com/huggingface/transformers}{https://github.com/huggingface/transformers}}.
We subsequently extract the embeddings using the flair library \cite{akbik:2019}.

\paragraph{Decontextualized}

To generate decontextualized embeddings, we again chose uncased BERT as starting point and the flair library for contextualization. For each token of interest, we collect the sentences it occurs in within the context datasets, generate contextualized embeddings for each of the sentences, and average them, as suggested by \citet{bommasani:2020}.

\end{document}